\documentclass[10pt,twocolumn,letterpaper]{article}

\usepackage{cvpr}
\usepackage{times}
\usepackage{epsfig}
\usepackage{graphicx}
\usepackage{amsmath}
\usepackage{amssymb}
\usepackage{caption}
\usepackage{subcaption}

\usepackage[breaklinks=true,bookmarks=false]{hyperref}

\cvprfinalcopy 

\def\cvprPaperID{****} 

\begin{document}

\title{S-NN: Stacked Neural Networks}

\author{
Milad Mohammadi\\
Stanford University\\
{\tt\small milad@cs.stanford.edu}
\and
Subhasis Das\\
Stanford University\\
{\tt\small subhasis@stanford.edu}
}

\maketitle

\begin{abstract}
It has been proven that transfer learning provides an easy way to achieve
state-of-the-art accuracies on several vision tasks by training a simple
classifier on top of features obtained from pre-trained neural networks. 

The goal of this project is to generate better features for transfer learning
from multiple publicly available pre-trained neural networks. To this end, we
propose a novel architecture called Stacked Neural Networks which leverages the
fast training time of transfer learning while simultaneously being much more
accurate. We show that using a stacked NN architecture can result in up to 8\%
improvements in accuracy over state-of-the-art techniques using only one
pre-trained network for transfer learning.

A second aim of this project is to make network fine-tuning retain the
generalizability of the base network to unseen tasks. To this end, we propose a
new technique called ``joint finetuning'' that is able to give accuracies
comparable to finetuning the same network individually over two datasets. We
also show that a jointly finetuned network generalizes better to unseen tasks
when compared to a network finetuned over a single task.
\end{abstract}


\section{Introduction}
\label{section:introduction}


Transfer learning is a general framework in which one trains a pre-trained
neural network to a new task on which the network was not trained. Amazingly,
Razavian et al.~\cite{razavian2014cnn} have shown that transfer learning on a
pre-trained neural network can outperform traditional hand-tuned approaches in
several tasks including coarse-grained detection, fine-grained detection and
attribute detection. In their work, Razavian et al. point out that ``It's all
about the features''. 

\begin{figure}
	\centering
	\includegraphics[width=0.9\columnwidth]{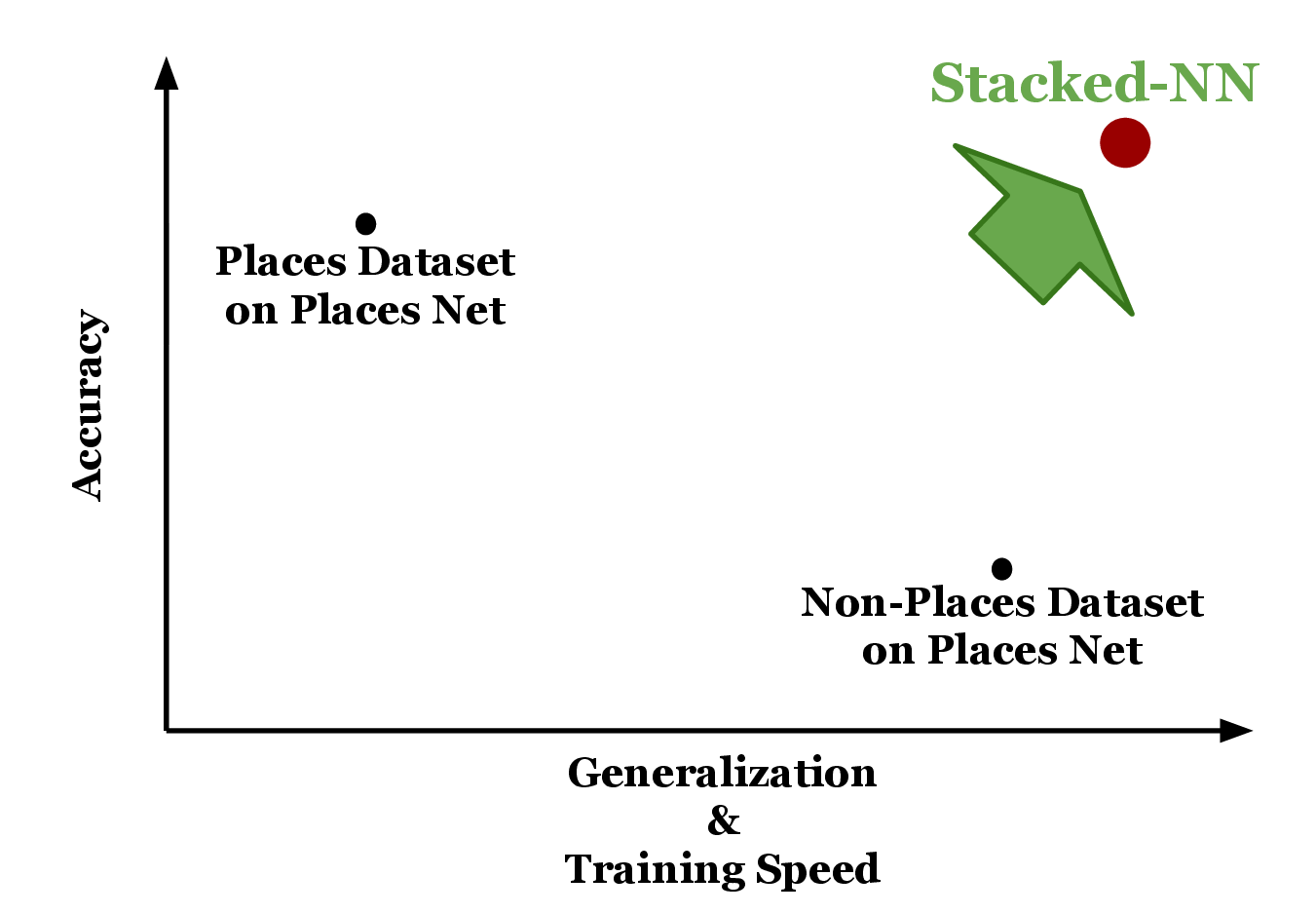}
    \caption{The conceptual state of the deep learning space presented as a
trade off between time consuming and task-specific neural network training
targeted specifically to specific datasets versus fast neural network training
through transfer learning to obtain reasonable performance from relatively
generalizable neural networks. The goal of this paper is to evaluate the
performance of a network architecture named \textit{Stacked Neural Networks
(S-NN)} to leverage the fast training speed of transfer learning while
considerably increasing the accuracy of transfer learning by \textit{generating
better features}.}
	\label{fig:insight}
\end{figure}

Figure~\ref{fig:insight} depicts the trade off between obtaining high
classification accuracy for deep networks highly trained for a particular
dataset for over a long time period and transfer learning to produce decent
classification accuracy over a short training time period. In order to reach
this target, this work is to present a novel method for leveraging higher
generalization accuracy from transfer learning.  Our aim is to \textit{find
better features} for vision datasets that are highly generalizable and fast to
train.

Since Razavian et al.'s work, several state-of-the-art pre-trained neural
networks have been made publicly available at Caffe Model Zoo~\cite{modelzoo}.
These include networks such as VGG~\cite{simonyan2014vgg},
GoogLeNet~\cite{princetonvision15}, Places~\cite{zhou2014places}, and
NIN~\cite{lin2013nin}.  Our initial studies suggested that these networks have
non-overlapping mis-classification behavior. This observation leads us to
believe that by combining the combination of these networks can improve the
classification by compensating for the shortcoming of other networks. What is
valuable to understand is whether there is a \textit{single} combination of
neural networks that is generalizable for all datasets or different combination
of networks provide optimal results for different datasets.

Before going any further, a short note about terminology. In this paper, we use the
term \textit{transfer learning} to mean training a SVM classifier on top of
features extracted from pre-trained networks without changing the networks. On
the other hand, by \textit{fine-tuning} we mean changing all the layers of a
network to better fit the given task.

Ensembling has been recognized as a simple way to boost the performance in a
vision task by averaging the scores of multiple networks together. However, in
some tasks such as Image-to-Sentence retrieval~\cite{mao2014mrnn}, a set of
\textit{features} is desirable instead of a score over some set of pre-defined
classes. Thus, we try to tackle the problem of \textit{generating better features}
rather than simply improving the transfer learning accuracy by ensembling.

In this work, we show that a combination of several network features by a novel
technique which we call \textit{stacking} offers better accuracy in many vision
tasks. We also evaluate various combinations of networks to find which network
combinations offer the best accuracy across the different datasets and whether
there is a single combination of networks that offers substantially higher
accuracy across the board.

We also evaluate the effect of using an ensemble of stacked networks rather than
a single stacked network in order to boost the performance of transfer learning
even further. We observe that ensembling can provide a substantial boost in
performance over and above that offered by stacking.

We also examine the effects of fine-tuning on the generalizability of the
features output by a network. We observe a significant drop in generalization
performance of a network once it has been fine-tuned to one specific task.
However, we show that \textit{joint finetuning} of a single network on two
different tasks can actually create a network that has accuracies close to the
individually fine-tuned networks. We also show that the features of a jointly
fine-tuned network have significantly higher generalizability on unseen tasks
than a fine-tuned network for a single task. However, we observe that this
jointly fine-tuned network still significantly underperforms the baseline of
using the pre-trained network features only. Section~\ref{section:joint} details
our experiments in this area.

\section{Neural Networks Set}
The network architectures and features used for this study are outlined below.

\textbf{VGG 16-layers and 19-layers (VGG16, VGG19)}: This is an architecture
proposed by Simonyan et al.~\cite{simonyan2014vgg} which uses a very deep
network (16 and 19 layers respectively) with smaller convolution filters of
3$\times$3 size to obtain state-of-the-art accuracies on the ImageNet 2014
Challenge. We use the \texttt{fc7} layer of the VGG network as the features
layer.

\textbf{GoogLeNet}: This is an architecture used by Szegedy et
al.~\cite{szegedy2014googlenet}, which uses several ``Inception'' modules to
create a deeper network with 22 layers while having much fewer parameters than
other networks such as VGG and AlexNet. We use \texttt{pool5/5x5\_s1} layer of
GoogLeNet as the features layer.

\textbf{Places}: This is a network created by Zhou et al.~\cite{zhou2014places}.
It has the same architecture as AlexNet~\cite{krizhevsky2012alexnet} but trained
on the Places dataset instead of ImageNet to enable better performance in
scene-centric tasks. We use the \texttt{fc7} layer of Places as the features
layer.

\textbf{Network In Network (NIN)}: This is a network architecture used by Lin et
al.~\cite{lin2013nin} which uses neural networks as the layer transfer function
instead of a convolution followed by a non-linearity. We use the \texttt{pool4}
layer as the features.

\section{Datasets}

%
Below, we describe the attributes of each of the datasets used for our study
evaluation.

\textbf{Caltech-UCSD Birds 200-2011~\cite{wah2011caltech}:} This is a dataset of 200
different species of birds. The dataset consists of 11,788 images.

\textbf{Caltech256~\cite{griffin2007caltech}:} This is a dataset of 256 object
categories containing 30,607 images. The dataset is collected from Google
images.

\textbf{Food-101~\cite{bossard14}:} This is a dataset of 101 distinct food categories
with 1,000 foods per category.

\textbf{LISA Traffic Sign Dataset~\cite{mogelmose2012vision}:} This dataset contains
7,855 annotations on 6,610 video frames captured on US roads. Each image is
labeled with the traffic signs visible on the images as well as the location of
the sign. It covers 47 of the US traffic signs.

\textbf{MIT scene~\cite{quattoni2009recognizing}:} This is an indoor scene dataset
with 15,620 images with 67 categories each of which containing at least 100
images.

\textbf{Oxford flowers~\cite{nilsback2008automated}:} This is a collection of 102
groups of flowers each with 40 to to 256 flowers.

\section{Methodology}


In this section, we formally define Stacked Neural Networks and discuss the
studies we conducted to construct a novel deep learning framework for improving
the state-of-the-art prediction accuracy of deep neural networks (NN). 

\subsection{Feature Stacking} 
\begin{figure}
	\centering
	\includegraphics[width=\columnwidth]{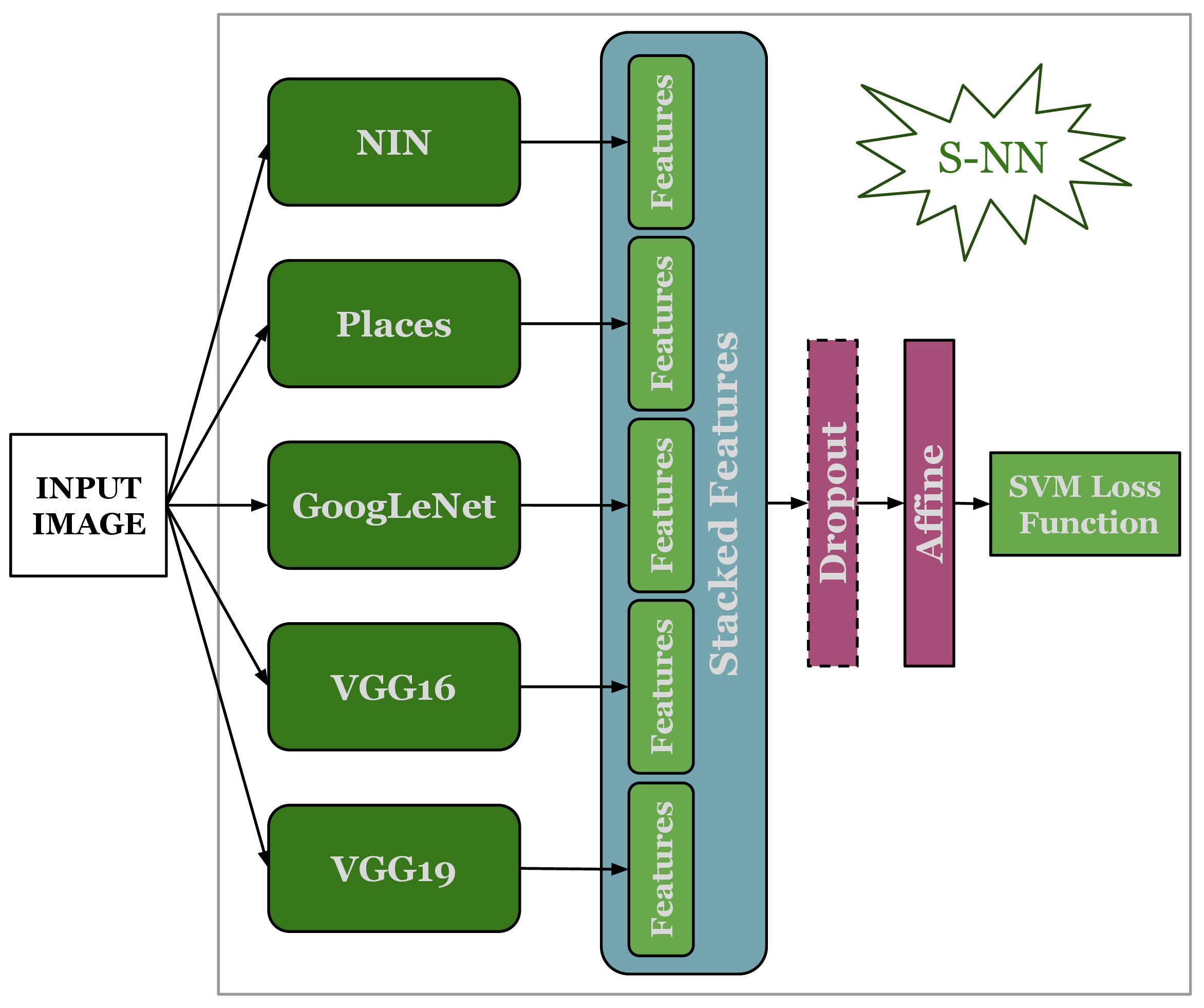}
    \caption{A stack of five publicly available neural network architectures.
The features generated from each network are combined into a unified feature
vector. This vector is used to classify the dropout and affine layers. We define
the combination of multiple networks a Stacked Neural Network (S-NN)}
	\label{fig:nn_stacked}
\end{figure}

Stacked Neural Networks (S-NN) is defined as a combination of publicly available
neural network architectures whose features are extracted at an intermediate
layer of the network, and then concatenated together to form a larger feature
set. Figure~\ref{fig:nn_stacked} illustrates this idea in detail. The
concatenated feature vector is used to train a classifier layer which consists
of an optional dropout layer, an affine layer and an SVM loss function. The
impact of the dropout layer will be discussed in detail in
Section~\ref{sec:result}.  While Figure~\ref{fig:nn_stacked} shows all five
convolutional neural networks (CNN) as members of the S-NN, any combination of
these CNN's is also considered a S-NN. For instance, \{GoogLeNet, VGG16\} and
\{NIN, Places, VGG19\} are examples of a 2-network and 3-network S-NN's. We will
evaluate the effect of different network combinations in
Section~\ref{sec:result} showing that S-NN's deliver higher classification
accuracy than single-network structures.

\subsection{Ensemble of S-NN's}
\begin{figure}
	\centering
	\includegraphics[width=\columnwidth]{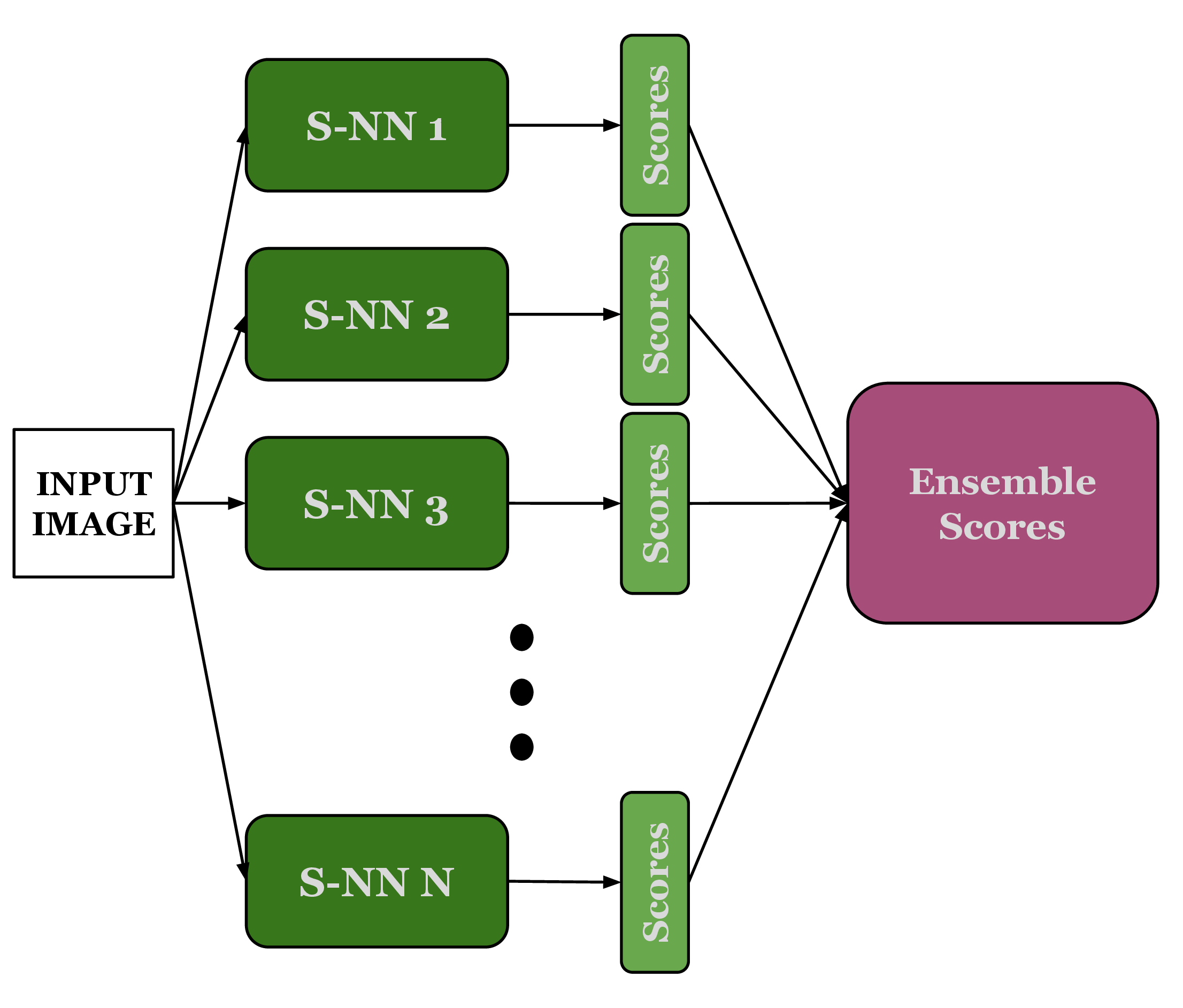}
    \caption{Scores generated from a group of S-NN's each containing a subset of
the networks illustrated in Figure~\ref{fig:nn_stacked} are combined to generate
the mean score of the ensemble.}
	\label{fig:nn_ensemble}
\end{figure}

It is shown in the literature~\cite{chenlearning} that an ensemble of multiple
independently trained networks can improve the prediction accuracy by reducing
the classification error rate. Each combination of S-NN's produces a new feature
vector and a new set of scores. To further improve the classification accuracy,
we studied the effect the ensemble mean of scores on the final network.
Figure~\ref{fig:nn_ensemble} shows a number of S-NN's whose scores are combined
into an ensemble score. While any arbitrary group of S-NN's may be used to
generate an ensemble score, we choose to compute this score by stacking a
S-NN with all its network combination subsets. For example, given a S-NN
containing three networks \{NIN, VGG19, GoogLeNet\}, we take the ensemble score
of all its network subsets: \{NIN\}, \{VGG19\}, \{GoogLeNet\}, \{NIN, VGG19\},
\{NIN, GoogLeNet\}, \{VGG19, GoogLeNet\}, \{NIN, VGG19\}, \{NIN, VGG19,
GoogLeNet\}. This method of forming ensembles allows us to compare the
performance of each S-NN combination against other S-NN's as discussed in detail
in Section~\ref{sec:result}, Figure~\ref{fig:combinations}.

\section{Results Discussion\label{sec:result}}

In this section we analyze the impact of S-NN and ensemble of S-NN's.
In doing so, we answer two key questions in this section:

\begin {enumerate}
    \item What is the impact of multiple networks on performance? In other
words, does more networks necessarily mean more performance? 
    \item What is the most generalizable S-NN architecture if we were to pick a
combination of these five NN's?
\end {enumerate}

\begin{figure}
	\centering
	\includegraphics[width=\columnwidth]{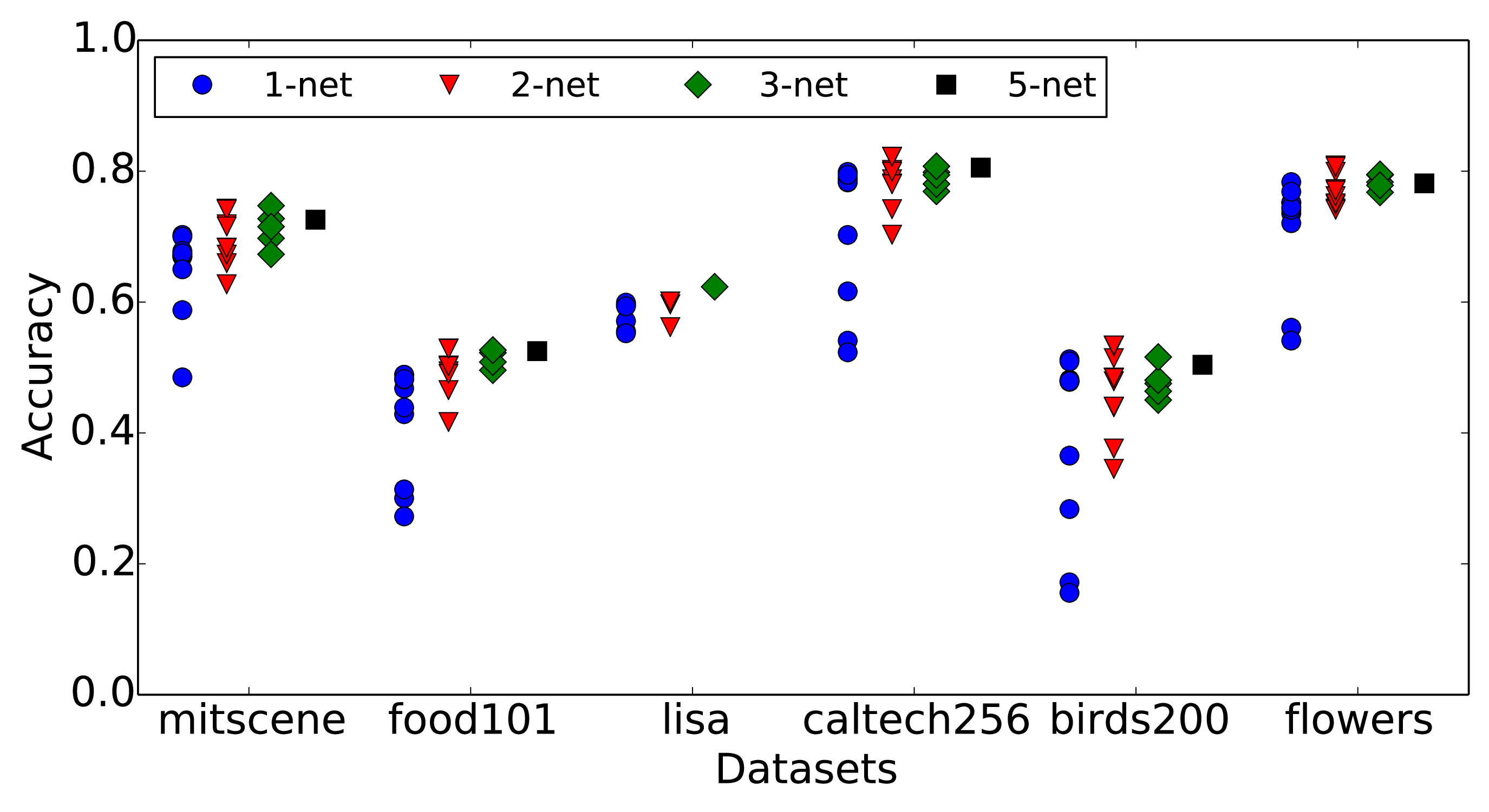}
    \caption{Best performing S-NN's over six different datasets and many
several different hyperparameters. S-NN's with 1,
2, 3, and 5 network are included. For each number of networks, all their
combinations are evaluated.}
	\label{fig:numnet}
\end{figure}

Figure~\ref{fig:numnet} is aimed to answer question (1). It shows our
\textit{best} experimental results on all combinations of 1, 2, 3, 5 S-NN's for
all datasets. Each S-NN combination was evaluated using 10's different
hyperparameter sweeps (i.e. learning rate, regularization factor, number of
epochs). The list of hyperparamters used in this work is included in
Table~\ref{tab:hyperparam}.  We evaluated the validation accuracy for single-network,
2-network, 3-network, and 5-network stack combinations.  All 2-network
experiments do not use the dropout layer while all 3-network experiments use the
dropout layer (see Figure~\ref{fig:nn_stacked}); we discovered dropout becomes
an important training element once the number of networks is larger than two.
For the case of five networks, however, we experimented the network performance
with and without the dropout layer.  Figure~\ref{fig:numnet} points out for
most networks, the case of 5 S-NN is favorable. It also indicates that the case
of 2 S-NN is strongly competitive with 5 S-NN.

\begin{table}[h]
\centering
\begin{tabular}{c | c}
\hline
Learning Rate & 1e-2, 5e-2, 1e-3, 2e-3\\
\hline
Regularization & 0.01, 0.1, 1, 10\\
\hline
Number of Epochs & 300, 400\\
\hline
Learning Rate Decay & 0.98\\
\hline
\end{tabular}
\caption{Hyperparameters used to profile the classification accuracy of all
network combinations used for S-NN experiments. The best classification
accuracy obtained from all these hyperparameter combinations is reported as a
data point in Figure~\ref{fig:numnet}.}
\label{tab:hyperparam}
\end{table}

\begin{figure}
	\centering
	\includegraphics[width=\columnwidth]{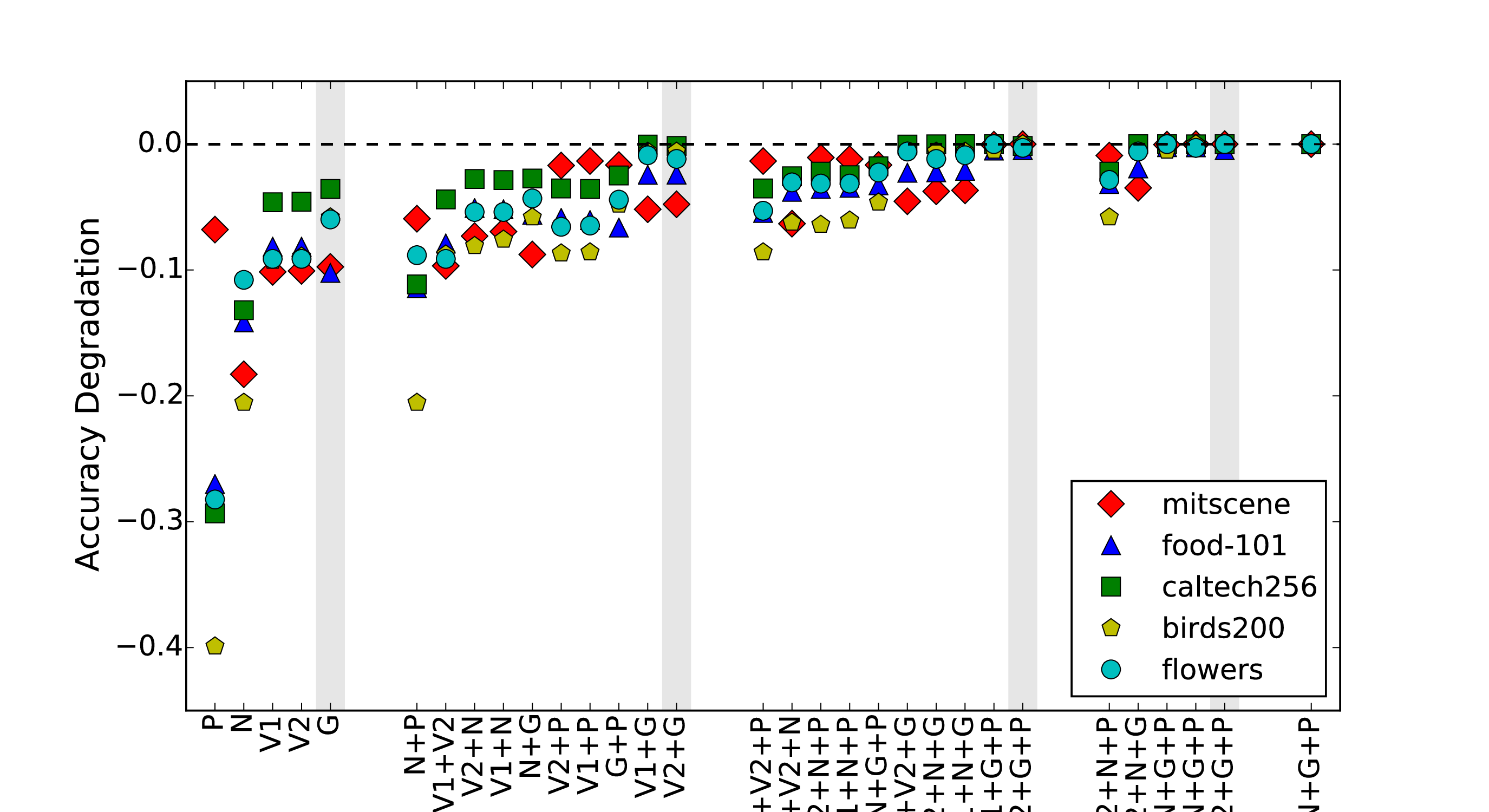}
    \caption{The horizontal axis shows all possible S-NN architectures. The
vertical axis shows the ensemble accuracy of each architecture with all
combinations of its subsets. For example, the subsets of \{NIN, VGG16\} used to
compute the ensemble accuracy are \{NIN, VGG16\}, \{VGG16\}, and \{NIN\}. The
network ensembles that deliver the highest generalization accuracy in the 1, 2,
3, and 4-network cases are highlighted in gray.}
	\label{fig:combinations}
\end{figure}

Figure~\ref{fig:combinations} is aimed to answer question (2). It shows all
possible network combinations and represents their accuracy degradation compared
to the best S-NN combination, which is 5-networks, for all datasets. This Figure
indicates for a single-network case, GoogLeNet is the most generalizable CNN as
it has the least mean and standard deviation in accuracy degradation. For the
2-network case, VGG16+GoogLeNet make the best network. This is a sensible choice
because (1) the two network are among the best publicly available networks, and
(2) they are constructed based on two different architectural
assumptions, making them relatively uncorrelated from the misclassification
behavior standpoint. For the 3-network case, VGG16+GoogLeNet+Places is the best
S-NN. For the 4-network case, VGG16+VGG19+GoogLeNet+Places form the best
classification choice indicating the poor generalization ability of the NIN CNN.

Notice the case with 5 networks has relatively negligible accuracy superiority
relative to the 2-network case, making two strong CNN's like GoogLeNet and VGG16
quite sufficient.

\begin{figure}
	\centering
	\includegraphics[width=\columnwidth]{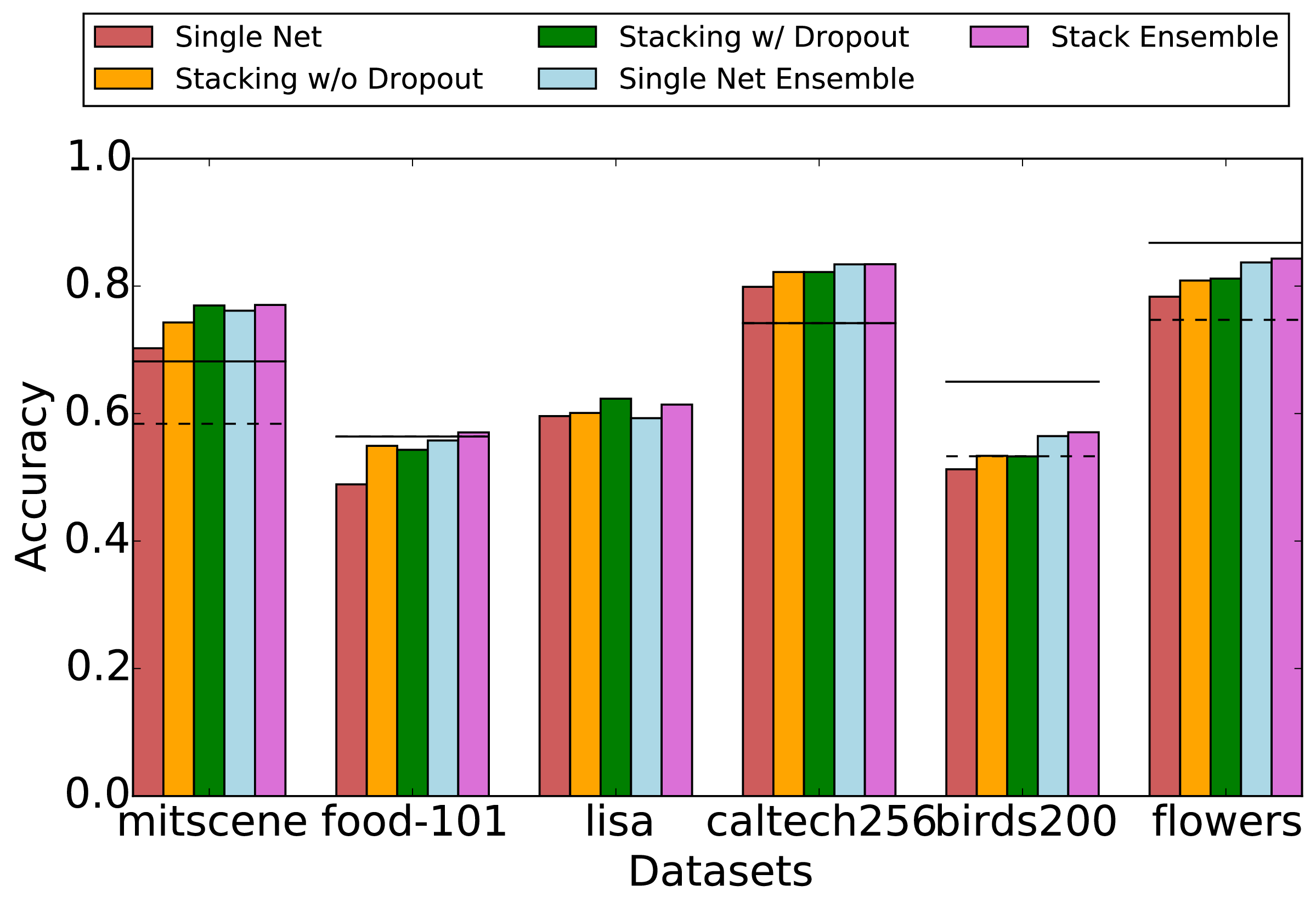}
    \caption{Best performance results generated across all tests done. The
dashed line represent the state-of-the-art without data augmentation and the
solid lines represent the state-of-the-art with data augmentation.}
	\label{fig:ensembles}
\end{figure}

Figure~\ref{fig:ensembles} focuses on extracting the best results obtained on
each dataset under different experimental settings. It compares the
single-network case with S-NN without dropout and S-NN with dropout
training. It further shows the best accuracies for the single-network
ensembles as well as the Stacked Ensemble model shown earlier. The
single-network ensemble refers to taking the mean of scores of individual
networks in order to construct the final score. Interestingly enough, these
results are more superior than the previous S-NN approach without score
ensembles. To further improve this performance, the Stack Ensemble mode also
includes the scores of S-NN architectures evaluated in Figure~\ref{fig:numnet}
in order to generate even more accurate classifications. 

The black lines in Figure~\ref{fig:ensembles} show the state-of-the-art accuracies in each given
dataset when data augmentation is applied and the dashed black lines show the
state-of-the-art accuracies without data augmentation~\cite {Nilsback06,
bossard14, wah2011caltech, griffin2007caltech, quattoni2009recognizing}.
Authors have not been able to identify any publications explicitly showing the
state-of-the-art classification results for the LISA dataset. While we have not
performed data augmentation as part of this study, we believe in doing so, we
can surpass the state-of-the-art performance results even on datasets our
results is below the solid line at the moment. 

\begin{figure}
\centering
    \begin{subfigure}{0.5\columnwidth}
        	\centering
        	\includegraphics[width=\columnwidth]{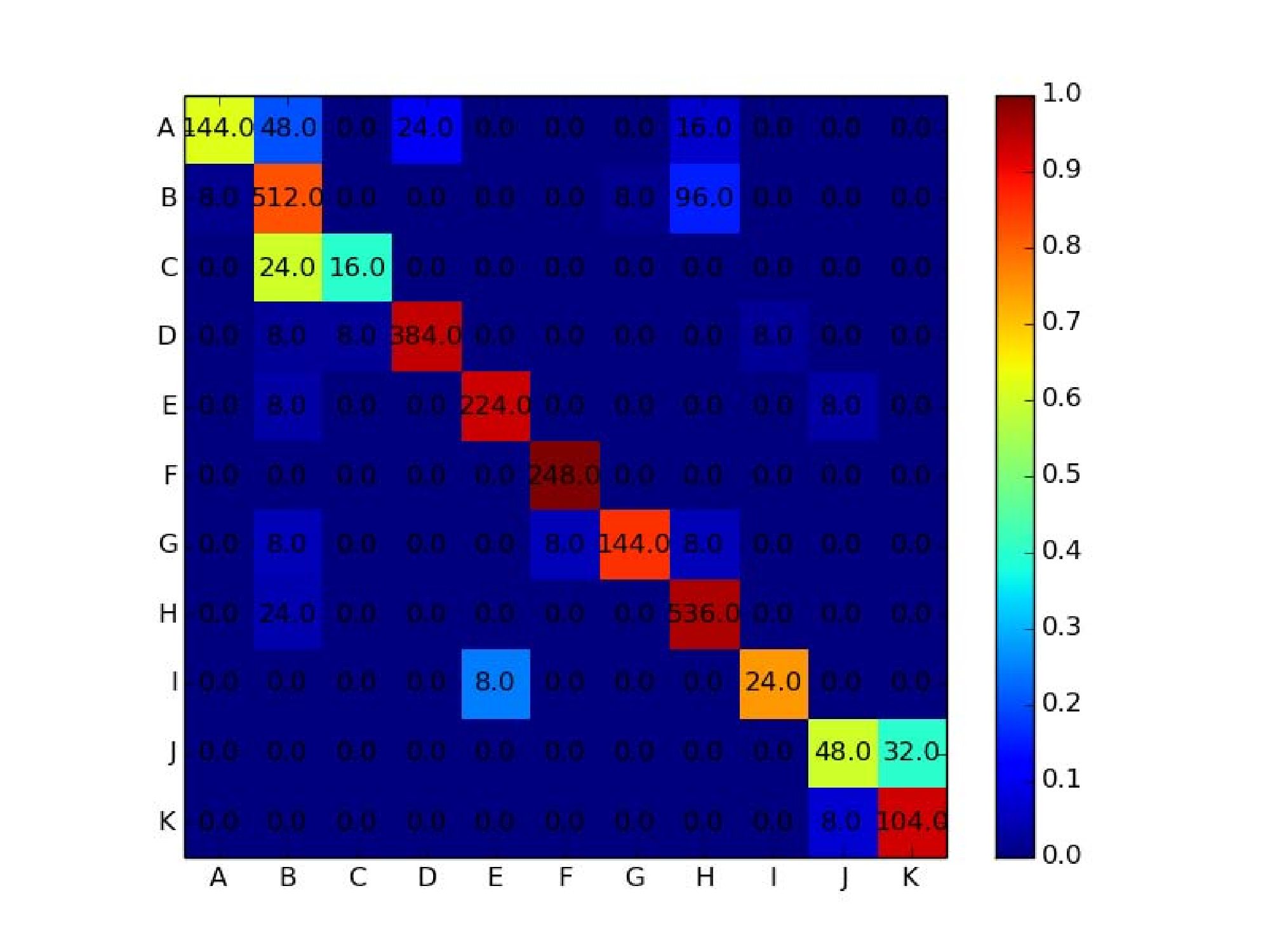}
            \caption{GoogLeNet CNN}
    \end{subfigure}%
    \begin{subfigure}{0.5\columnwidth}
        	\centering
        	\includegraphics[width=\columnwidth]{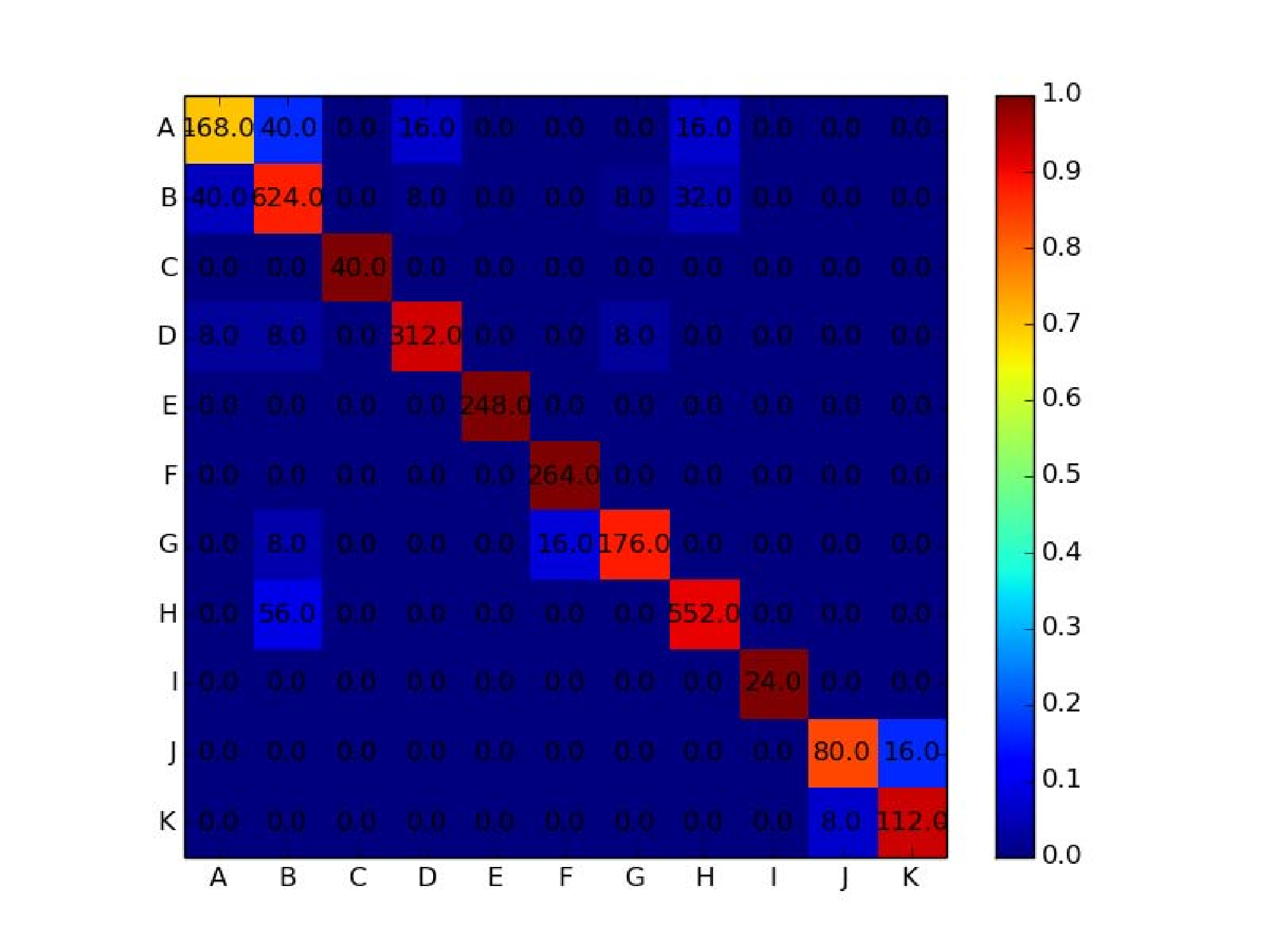}
            \caption{Places CNN}
    \end{subfigure}
    \begin{subfigure}{0.5\columnwidth}
        \centering
        \includegraphics[width=\columnwidth]{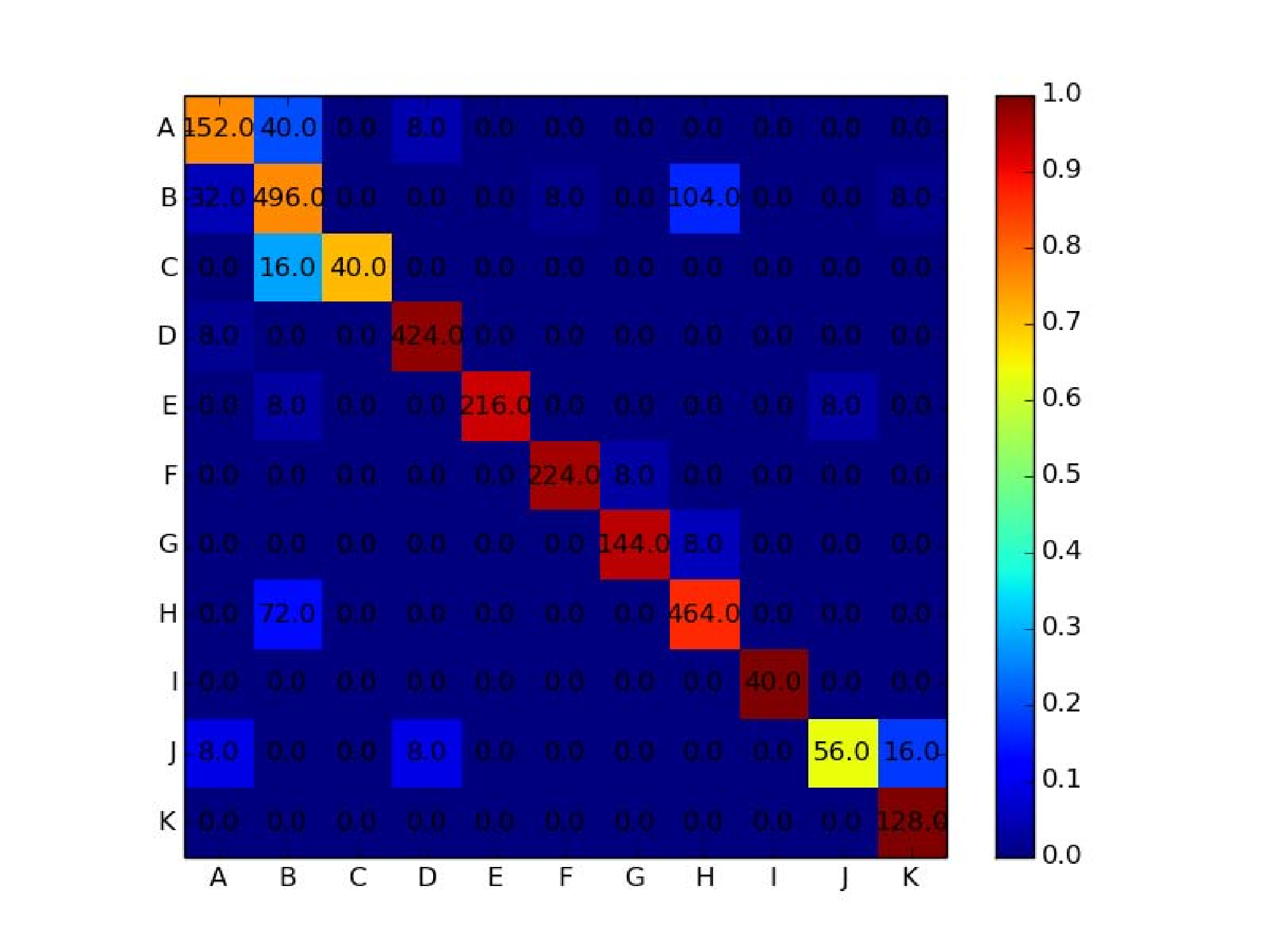}
        \caption{VGG16 CNN}
    \end{subfigure}%
    \begin{subfigure}{0.5\columnwidth}
        \centering
        \includegraphics[width=\columnwidth]{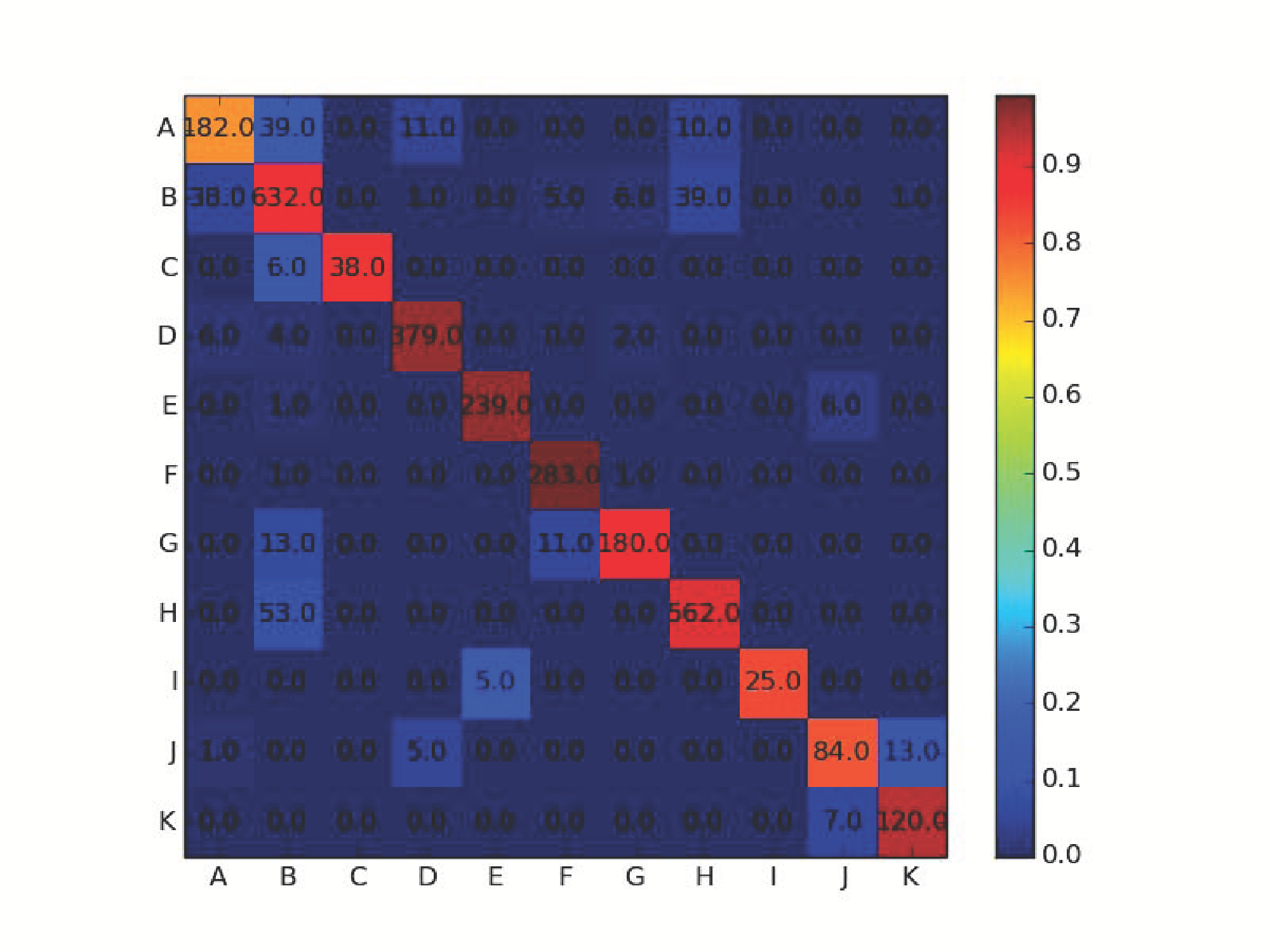}
        \caption{VGG16+GoogLeNet+Places CNN's}
    \end{subfigure}
    \caption{Confusion matrix of the MIT Scene dataset run on three independent
CNN's along with their S-NN model. This dataset has 67 classes. To avoid
clutter, these matrices only consider 11 of these classes. The value in each
cell corresponds to the number of instances an image in actual class X is
classified as Y. The main diagonal represents correct classification.}
    \label{fig:conf_mats}
\end{figure}

Figure~\ref{fig:conf_mats} shows the confusion matrices for the standalone
classification performance of the GoogLeNet, Places, and VGG16 CNN's when
running the MIT Scene dataset. To avoid clutter, this Figure illustrates 11
classes. Here, we draw a number of interesting insights from these results when
comparing similar confusion cells in different matrices:

\begin {itemize}
    \item The BH misclassification happens 104 times in VGG16 and 96 times in
GoogLeNet while only 32 times in Places. The combined S-NN only has 39
misclassifications. This shows the error correction power networks features in
compensating for each other's weakness. This effect can also be observed in DI
where the misclassification for the S-NN is zero while GoogLeNet
misclassifies 8 images in this cell.
    \item  The IE misclassification happens 8 times in GoogLeNet and 0 times in
Places and VGG16. Despite the presence of VGG16 and Places network, the S-NN
confuses 5 images to this cell. This shows despite the presence of other
networks to completely eliminate this type of image misclassification in S-NN,
the score from GoogLeNet dominates the overall outcome more often. In
Section~\ref{sec:weighted_snn}, a potential solution to this problem is
discussed.
    \item The FB misclassification never happens to single-net cases. However,
the stack of features shows 1 misclassification in this cell. While a negligible
misclassification case, this shows the stack of NN features can have some
adversarial effect in the final outcome. In Section~\ref{sec:weighted_snn}, a
potential solution to this problem is discussed.
\end {itemize}

\section{Joint Training of Multiple Tasks}
\label{section:joint}

As part of this work, we also looked at how the generalization of network
features suffers as a result of finetuning a network on a given task. To perform
these experiments, we considered three tasks from different domains: Food-101
(task A), MIT Scene (task B), and Caltech256 (task C). In all of these
experiments, we used the VGG16 network. We describe our experiments and
observations below.

Recall from Section~\ref{section:introduction} that by \textit{fine-tuning} we
mean actually changing the network parameters, while by \textit{transfer
learning} we mean only using an SVM layer on top of a pre-trained network.

\subsection{Generalization Loss by Fine-Tuning}
In our first experiment, we investigated the generalization loss by network
finetuning and ways to avoid this problem.

To show loss of generalizability by fine-tuning, we first fine-tuned VGG16 on
task A to create the network VGG16A, and independently on task B to create a
network VGG16B. Subsequently, we evaluated the transfer learning accuracy of
task B on VGG16A. We compared this with the accuracy of VGG16B on task B. The
results are summarized in Table~\ref{table:jointresults}.

This comparison shows that indeed fine-tuning VGG16 on task A reduced the
transfer-learning accuracy on task B drastically (43.0\% compared to 72.2\%).
Thus, an interesting question is whether it is possible to fine-tune a single
network to give accuracies similar to the individually fine-tuned networks on
\textit{both} the tasks A and B?

Our experiments show that the answer to this question is ``Yes''. Below we
describe the methodology for achieving this result.

\begin{figure}
    \centering
    \includegraphics[width=0.8\columnwidth]{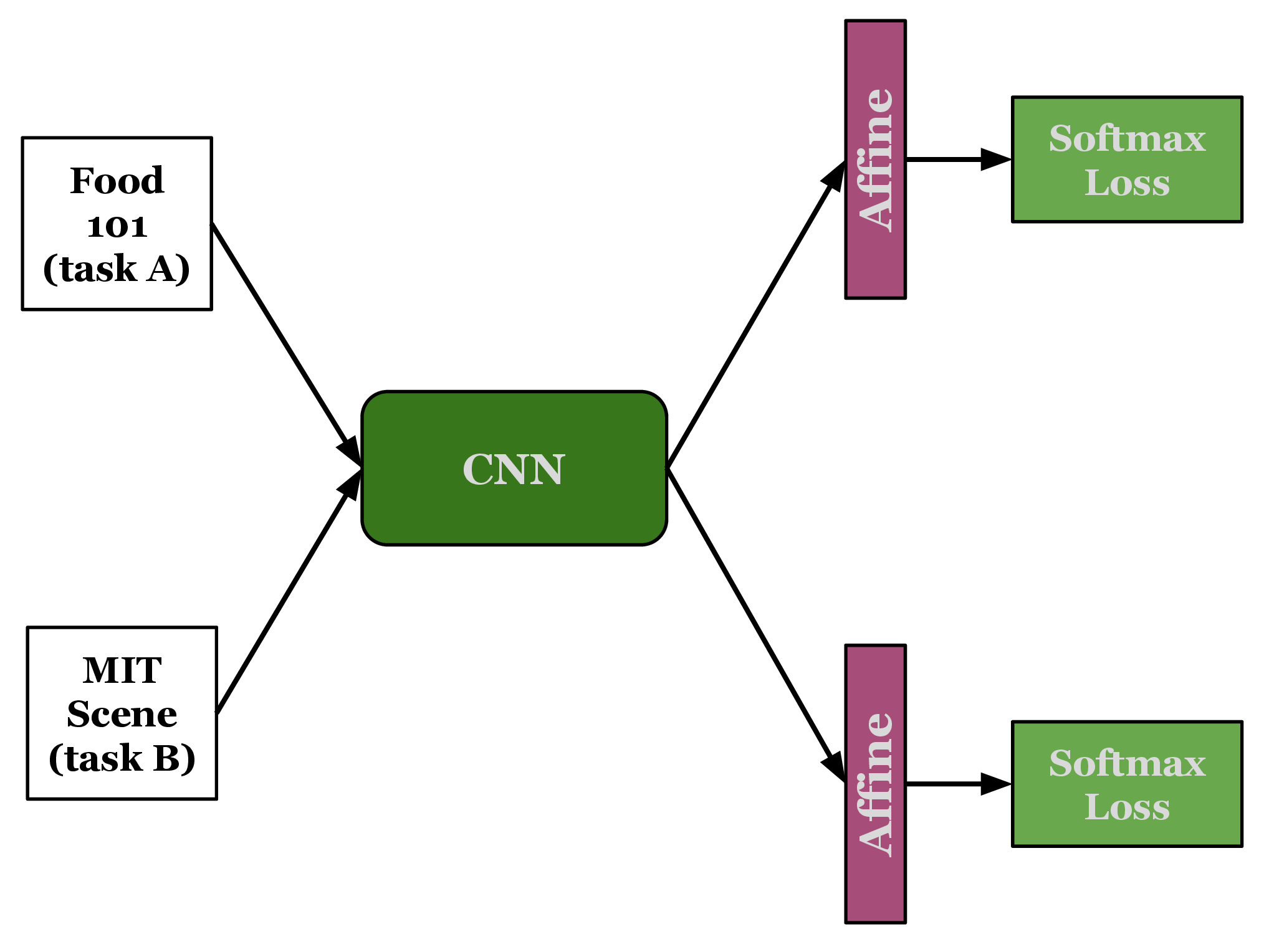}
    \caption{One neural network trained on multiple tasks with multiple
classifier layers, one per task.}
    \label{fig:joint}
\end{figure}

The network architecture is shown in Figure~\ref{fig:joint}. We trained a single
network with the concatenation of multiple dataset inputs in each minibatch.
Each minibatch consisted of 4 images, 2 of which were from task A and 2 from
task B. The output features from the network are then fed into two independent
linear classifier layers, one for each of the tasks. The final loss is taken to
be the sum of softmax losses from the two classifier layers. Let us denote this
jointly trained network by VGG16AB.

Table~\ref{table:jointresults} shows the resulting accuracy from this approach.
It can be seen that the accuracy of VGG16AB on task A (68.0\%) is close to the
accuracy of VGG16A (69.6\%) while the accuracy on task B (70.6\%) is close to
VGG16B (72.2\%). Thus, it can be inferred that it is indeed possible to gain the
best of both VGG16A and VGG16B in a single network.

\begin{table}[h]
\centering
\begin{tabular}{c | c}
\hline
{\color{red}task A} on {\color{red}VGG16A} & 69.6\%\\
\hline
{\color{blue}task B} on {\color{blue}VGG16B} & 72.2\%\\
\hline
{\color{blue}task B} on {\color{red}VGG16A} & 43.0\%\\
\hline
{\color{red}task A} on {\color{green}VGG16AB} & 68.0\%\\
\hline
{\color{blue}task B} on {\color{green}VGG16AB} & 70.6\%\\
\hline
\end{tabular}
\caption{Performance of {\color{red}task A} and {\color{blue}task B} on
{\color{red}VGG16A}, {\color{blue}VGG16B}, and {\color{green}VGG16AB}}
\label{table:jointresults}
\end{table}

\subsection{Generalization of Jointly-Tuned networks}
In the previous section, we saw that jointly tuning a network on two tasks can
give accuracies comparable to individual finetuning. We also saw that individual
finetuning destroys a lot of the generalization capabilities of the original
network. These two facts bring up the next question: what is the generalization
capability of a jointly tuned network?

To answer this question, we took a third task C (Caltech256 in our experiments),
and evaluated the transfer learning accuracy of task C over VGG16, VGG16A and
VGG16AB.  The results are summarized in Table~\ref{table:genjoint}.

\begin{table}[h]
\centering
\begin{tabular}{c | c}
\hline
{\color{cyan}task C} on {\color{black}VGG16} & 75.0\%\\
\hline
{\color{cyan}task C} on {\color{red}VGG16A} & 54.8\%\\
\hline
{\color{cyan}task C} on {\color{green}VGG16AB} & 61.3\%\\
\hline
\end{tabular}
\caption{Performance of {\color{cyan}task C} on VGG16,
{\color{red}VGG16A} and {\color{green}VGG16AB}}
\label{table:genjoint}
\end{table}

It can be seen that, indeed, the generalization capabilities of a jointly
finetuned network are higher. Although the jointly finetuned network still
does not reach the transfer learning capability of the baseline network alone,
it does succeed in mitigating a lot of the degradation coming from individual
finetuning. We suspect that the reason behind this phenomenon is that the
jointly-finetuned network is forced to finetune towards more ``general''
properties of the tasks rather than being extremely specific to one
particular task.

From these two experiments, we can conclude that joint finetuning is a promising
direction to look into for finetuning networks without losing their
generalization capabilities.

\section{Future Direction}

%

\subsection{Parallel Training of Networks}
As shown in~\cite{yosinski2014transferable}, training neural networks at deeper
layers can improve the classification accuracy.  While the S-NN's, explained in
the previous sections, are targeted toward agile training and high
generalization accuracy, it is reasonable to hypothesize if they were trained
together, their collective classification error would drop. So, instead of using
the data for when each of these networks were pre-trained independently, we
allow the backward propagation training method to broadcast the classifier
gradients to all networks.  Since all networks iterate through the same set of
gradients in parallel, they all influence each other's weight and bias values.
Due to the limited computation capability available to us, we have been unable
to generate results for this step. We will continue working on this scheme via
once we have access to more powerful GPU machines.

\subsection{Weighted S-NN\label{sec:weighted_snn}}
Our analysis of the confusion matrices on the available datasets made us realize
while the combination of network features in S-NN helps reduce errors, it is
also the case that some networks introduce excessive confusion to the error rate
of S-NN. To reduce such adversarial impacts of S-NN, we plan to combine features
by applying weights to each network feature. The weight coefficients will be
dependent on the prediction accuracy of each network in classifying a given
dataset. For instance, if GoogLetNet CNN shows weaker classification performance
relative to the Places CNN, the relative contribution of GoogLeNet features will
be reduced. To do so, the feature vector of each network is multiplied by a
scalar value in [0, 1]. This value is computed by dividing the classification
accuracy of each NN by the accuracy of the network with the best result. For
instance, assume a S-NN consisting of \{GoogLeNet, VGG16\}. If GoogLeNet and
VGG16 have individual classification accuracy of 0.3 and 0.6 respectively, the
GoogLeNet and VGG features are multiplied by 0.5 and 1 respectively before
stacking their features.

\subsection{Data Augmentation}
Data augmentation has improved classification accuracy via diversifying NN
features. Literatures \cite{Nilsback06, bossard14, wah2011caltech} show
substantial improvement in the MIT Scene, CUB 200, and Oxford Flowers datasets
when their networks are trained using data augmentation. While we have been
short in time to try this technique, we believe it will substantially boost our
prediction accuracy on most datasets if not all.

\subsection{Parallel Wimpy Networks}
Inspired by the notion of ensembles presented by Hinton et
al.~\cite{darkKnowledge}, this study proves combining multiple powerful networks
leads to more substantial performance gains. It also proves training a
single network on multiple datasets can deliver better generalization accuracy.
The next milestone we would like to tackle is to evaluate the possibility of
building numerous small, fast-to-train networks trained on multiple datasets and
stacked as S-NN's. We call them a stack of \textit{wimpy neural networks}. Such
a technique is interesting to us from two fronts.  First is to find if multiple
wimpy S-NN's can do as well as (or better than) a powerful network like VGG19.
Second is to find if this architecture can help reduce computation overhead
demand of recent deep neural networks, such as VGG19 by enabling a much more
parallelizable network architecture with the ability to be conveniently
offloaded onto multiple computation units (i.e. CPU's or GPU's).

\section{Conclusion}
In this work, we presented Stacked Neural Networks, a novel technique in
extracting higher generalization accuracy from the state-of-the-art neural
networks in the public domain. We evaluated various NN stack combinations and
discovered that while a five-CNN stack delivers the best accuracy, the stack of
two CNN's can deliver similar accuracy gains while consuming much less
computation power. We also presented the classification accuracy improvements of
generating the ensemble of S-NN's.  The combination of these techniques enabled
us to boost the classification accuracy beyond the state-of-the-art results
presented in previous literature.

Furthermore, we evaluated the effect of training multiple datasets on one
network. Interestingly enough, we concluded that it is possible to jointly
finetune a single network over multiple datasets and still obtain accuracies
that are almost similar to individual finetuning of the networks. We also show
that these jointly finetuned networks have better generalization capabilities
than individually finetuned variants.

S-NN proves the presence of fruitful impact in fostering collaborative neural
network classification to improve generalization accuracy in transfer learning.

{\small
\bibliographystyle{ieee}
\bibliography{egbib}
}

\end{document}